\documentclass[sigconf]{acmart}
\usepackage{algorithm, algorithmicx}
\usepackage[noend]{algpseudocode}
\usepackage{adjustbox}
\usepackage{multirow}
\usepackage{enumitem}
\usepackage{balance}

\AtBeginDocument{%
  }

\copyrightyear{2024}
\acmYear{2024}
\setcopyright{acmlicensed}

\acmConference[CIKM '24] {Proceedings of the 33rd ACM International Conference on Information and Knowledge Management}{October 21--25, 2024}{Boise, ID, USA.}
\acmBooktitle{Proceedings of the 33rd ACM International Conference on Information and Knowledge Management (CIKM '24), October 21--25, 2024, Boise, ID, USA}
\acmISBN{979-8-4007-0436-9/24/10}
\acmDOI{10.1145/3627673.3680013}

\begin{document}

\title{An End-to-End Reinforcement Learning Based Approach for Micro-View Order-Dispatching in Ride-Hailing}


\author{Xinlang Yue}
\orcid{0009-0003-8228-5256}
\affiliation{%
  \institution{Didi Chuxing}
  \city{Beijing}
  \country{China}
}
\email{xlyue92@outlook.com}

\author{Yiran Liu}
\orcid{0009-0009-3819-6743}
\affiliation{%
  \institution{Didi Chuxing}
  \city{Beijing}
  \country{China}
}
\email{yiran.l@outlook.com}

\author{Fangzhou Shi}
\orcid{0009-0000-4210-712X}
\affiliation{%
  \institution{Didi Chuxing}
  \city{Beijing}
  \country{China}
}
\email{arkshifangzhou}
\email{@didiglobal.com}

\author{Sihong Luo}
\orcid{0009-0006-6450-0516}
\affiliation{%
  \institution{Didi Chuxing}
  \city{Beijing}
  \country{China}
}
\email{arpharluo@didiglobal.com}

\author{Chen Zhong}
\orcid{0009-0001-7503-3064}
\affiliation{%
  \institution{Didi Chuxing}
  \city{Beijing}
  \country{China}
}
\email{zhongchen@didiglobal.com}

\author{Min Lu}
\orcid{0009-0001-2238-5365}
\affiliation{%
  \institution{Didi Chuxing}
  \city{Beijing}
  \country{China}
}
\email{lumin@didiglobal.com}

\author{Zhe Xu}
\orcid{0000-0002-3902-2595}
\affiliation{%
  \institution{Didi Chuxing}
  \city{Beijing}
  \country{China}
}
\email{xuzhejesse@didiglobal.com}

\renewcommand{\shortauthors}{Xinlang Yue et al.}

\begin{abstract}
  Assigning orders to drivers under localized spatiotemporal context (micro-view order-dispatching) is a major task in Didi, as it influences ride-hailing service experience. Existing industrial solutions mainly follow a two-stage pattern that incorporate heuristic or learning-based algorithms with naive combinatorial methods, tackling the uncertainty of both sides’ behaviors, including emerging timings, spatial relationships, and travel duration, etc. In this paper, we propose a one-stage end-to-end reinforcement learning based order-dispatching approach that solves behavior prediction and combinatorial optimization uniformly in a sequential decision-making manner. Specifically, we employ a two-layer Markov Decision Process framework to model this problem, and present \underline{D}eep \underline{D}ouble \underline{S}calable \underline{N}etwork (D2SN), an encoder-decoder structure network to generate order-driver assignments directly and stop assignments accordingly. Besides, by leveraging contextual dynamics, our approach can adapt to the behavioral patterns for better performance. Extensive experiments on Didi’s real-world benchmarks justify that the proposed approach significantly outperforms competitive baselines in optimizing matching efficiency and user experience tasks. In addition, we evaluate the deployment outline and discuss the gains and experiences obtained during the deployment tests from the view of large-scale engineering implementation.
\end{abstract}


%

\maketitle
\section{Introduction}\label{sec:intro}
Order-dispatching – assigning passengers' orders to available drivers in real-time – is the key process in ride-hailing platforms, influencing service experience of both drivers and passengers. There are primarily two research scopes on this topic. The macro-view scope \cite{xu2018large, lin2018efficient, tang2019deep, tang2021value} focuses on long-term (several hours to a day) and city-level efficiency optimization. The other scope attends to the optimization under localized spatiotemporal scenarios with high stochasticity, i.e., \underline{mic}ro-view \underline{o}rder-\underline{d}ispatching (MICOD). This problem centers on matching unspecified number of drivers and orders in each decision window of fixed seconds (a batch) \cite{qin2022reinforcement}, optimizing goals (measured by driver income, pickup distance, etc.) over multiple batches within a localized area (typically 10 minutes over tens of geo-grids, i.e., geo-fence).
\begin{figure}[t]
    \setlength{\belowcaptionskip}{-7mm}
    \centering
    \includegraphics[width=\linewidth]{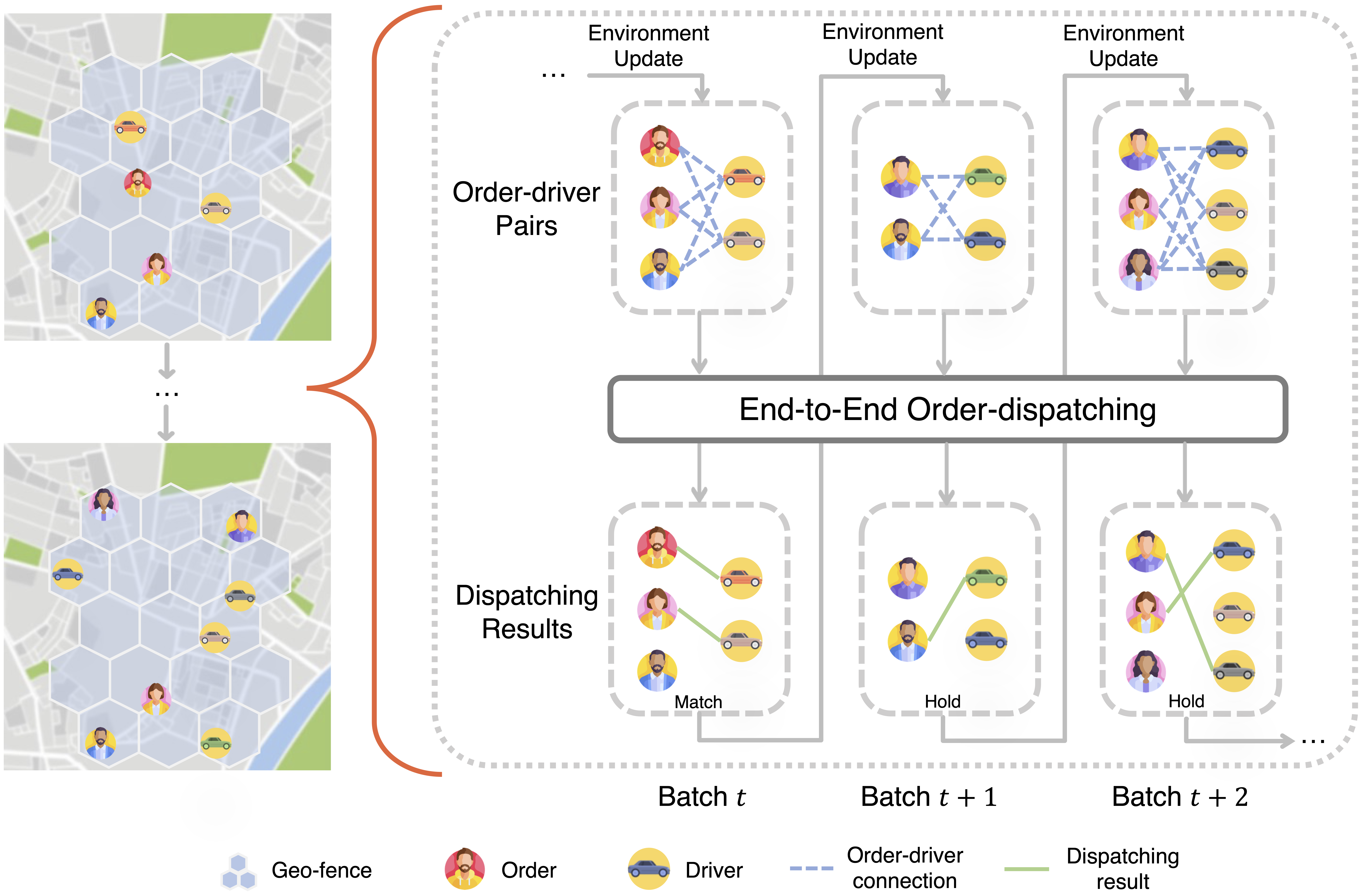}
    \caption{Micro-view order-dispatching with the proposed end-to-end framework.}\label{titleimage}
\end{figure}

Given its online nature, the major challenge of MICOD arises from its bilateral dynamics, in which both the number and the contextual attributes of drivers and orders remain unknown in upcoming batches. Therefore, the problem can be regarded as a sequential decision-making problem. Each decision within a batch entails solving a combinatorial optimization (CO) task on order-driver (o-d) assignment, while the goal is to maximize the global gain over a sequence of batches. Given the changing bipartite size in each batch, the problem is demanding with the decision space being combinatorial and boundlessly large.

In the MICOD context, the mainstream industrial methodologies follow a two-stage pattern of "holding + dispatching". The dispatching part includes general CO approaches, such as the greedy method \cite{kalyanasundaram1993online}, the Hungarian algorithm \cite{kuhn1955hungarian} and the stable matching method \cite{gale1962college}.
While naive methods can manage single batch optimization almost perfectly, it is worth noting that myopic optimization of each CO problem does not guarantee the global gain across multiple batches. For example, a passenger may wait a few more batches for a driver who is 0.5 miles away instead of matching
with a driver at a distance of 3 miles immediately. Therefore, it is necessary to take future into account and hold less appropriate o-d pairs for future dispatching. 
In Didi's practice, we predict the optimal matching moment \cite{shiryaev2007optimal} with deep learning (DL) models rather than dispatching them immediately. This strategy is called spatiotemporal hold (StH) and is widely applied in Didi’s online environment. 

In addition, some recent works focus on predicting matching frequency or altering the volume of o-d pairs in each batch with a deep reinforcement learning (DRL) framework.
For example, \cite{qin2021optimizing} attempts to learn a centralized delay-matching policy. This policy dynamically adjusts the matching frequency of each pre-partitioned area based on demand and supply (D\&S) information, aiming to optimize the overall matching efficiency. \cite{wang2019adaptive} introduces a restricted batch-splitting policy integrated with the Hungarian algorithm. In spite of above efforts, these two-stage methods regard bipartite matching as a part of the environment rather than actions, sidestepping bilateral uncertainty of the problem. Moreover, this pattern bears the natural inconsistency between agent holding actions and overall optimization goal of the entire dispatching process.

In this paper, we manage to transform to a one-stage "DRL for dispatching" pattern with implicit behavior prediction, addressing bilateral uncertainty of MICOD by directly generating o-d assignments, as illustrated in Figure \ref{titleimage}. Furthermore, we present an end-to-end DRL approach to avoid inconsistency between optimization objectives and decision actions by fully leveraging contextual and environmental information. To achieve this, we introduce a two-layer Markov Decision Process (MDP) framework to entirely model MICOD, and propose \underline{D}eep \underline{D}ouble \underline{S}calable \underline{N}etwork (D2SN), a novel deep model, to output assignment decisions auto-regressively \cite{chen2018pixelsnail} to adapt to changing decision space in each batch.

The main contributions of this paper are summarized as follows:
\begin{itemize}[leftmargin=*, topsep=0.5mm]
    \item We propose a one-stage DRL approach to uniformly solve behavior prediction, sequential decision-making and combinatorial optimization in the entire MICOD, addressing the bilateral uncertainty with consistent optimization goals.
    
    \item Specifically, we introduce a two-layer MDP framework to model MICOD and propose D2SN to generate o-d assignments directly while accommodating aforementioned challenges.

    \item Given real-world benchmarks and a calibrated simulator by Didi, in-depth experiments justify that D2SN outperforms CO and two-stage DRL baselines with notable improvements in terms of efficiency and user experience optimization.
\end{itemize}
\section{Related Works}\label{sec:rw}
\textbf{Reinforcement learning (RL) in ride-hailing}. 
With the prevalence of RL, abundant researches formulate large-scale ride-hailing problems in an MDP setting and attempt to solve them in a value-based way. It is intuitive to model each driver as an agent \cite{wang2018deep, sun2022optimizing, holler2019deep}, such that the scalability in action space can be easily handled, usually by learning a tabular or state value function \cite{xu2018large, tong2021combinatorial, han2022real}.
In addition, there are some multi-agent RL (MARL) approaches \cite{li2019efficient, jin2019coride, zhou2019multi}.
\cite{li2019efficient} utilizes the mean-field MARL framework to model the interaction among neighboring agents.
\cite{zhou2019multi} designs a Kullback-Leibler divergence regularizer to deal with the discrepancy of the D\&S distributions.
\cite{jin2019coride} models the problem in a hierarchical setting, regarding hex cell as worker agents and groups of hex cells as manager agents.
However, most existing works assume driver homogeneity with shared policies and ignore the contextual diversity.

\noindent \textbf{Online matching in order-dispatching.} Researches on online matching \cite{karp1990optimal} have been widely conducted in industries, such as online allocation \cite{ho2012online, zeng2023deep}, crowd-sourcing \cite{zeng2018latency, ma2021hierarchical}, etc.
The MICOD scenario is special for its changing bilateral dynamics \cite{hu2022dynamic} and the batching operating manner.
For the holding part in two-stage methods,
some works \cite{qin2021optimizing, wang2019adaptive, ke2020learning} consider drivers and orders eligibility of entering the bipartite matching and also investigate batched window optimization using RL methods to achieve long-term gains. \cite{yang2020optimizing} attempts to jointly optimize matching radius and batch size to adjust candidates in each batch. Basically, these approaches treat the dispatching process as a part of the environment changes.
\section{Methodology}\label{sec:problem} 
\subsection{MICOD Formulation}\label{micod}
A typical MICOD problem in Didi must comply with:

\begin{itemize}[leftmargin=*]
    \item Micro-view: the spatiotemporal range of MICOD is restricted to a brief time period (usually 10 minutes) and a geo-fence (around 30 square kilometers, consists of 40-50 geo-grids) \cite{tang2019deep}.
    \item Dynamic contextual patterns: Both drivers and orders may emerge or go offline following their own or joint behavioral patterns before or during service.
    \item Batch mode \cite{qin2022reinforcement}: MICOD performs order-dispatching in a batch mode. In each batch (time window), the size-unspecified o-d assignment can be formulated as a CO problem. 
\end{itemize}

With above restrictions, the objective is to maximize the cumulative gain over the entire period of all batches. Besides, the micro-view and dynamic properties prompt the exploration of the issue by thoroughly leveraging contextual information for each o-d assignment. Note that we restrict to one-to-one assignment (one driver can serve at most one order) in MICOD. Ride-pooling, serial-assign, etc. are not considered due to their low proportion in Didi's online environment. We leave them for future research.

\subsection{Two-layer MDP Framework}
With the formulation in Sec. \ref{micod}, we model the problem as an MDP process. The order-dispatching system is the agent, and the action is to select o-d pairs as dispatching results. Due to the dynamic contexts in MICOD, the action space is changing in each batch, making regular MDP transitions \cite{qin2021optimizing, yang2020optimizing, wang2019adaptive} hard to model. Instead, we seek to decompose the MDP setting into a two-layer MDP framework \cite{metz2017discrete} with the same agent.

\begin{figure}[t!]
    \setlength{\belowcaptionskip}{-4mm}
    \centering
    \includegraphics[width=\linewidth]{{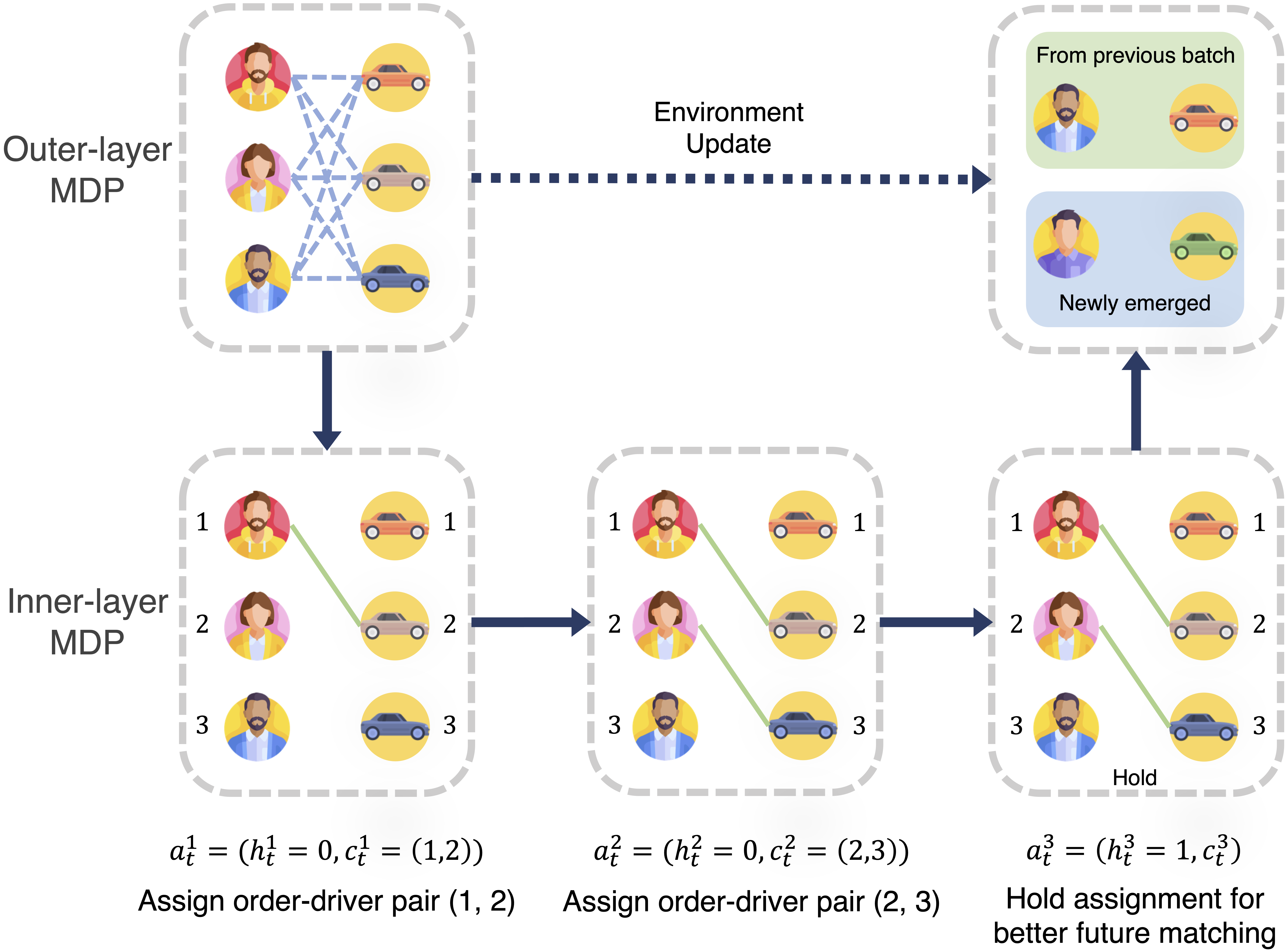}}
    \caption{Order-dispatching process under the two-layer MDP framework. The upper shows the outer-layer state transition. The lower shows the inner-layer sub-state transition.}\label{twolayermdp}
\end{figure}

Figure \ref{twolayermdp} illustrates the dispatching process under the two-layer MDP setting. The agent first collects all available o-d pairs and global information in a batch as the outer-layer MDP state, which is also the initial sub-state of the inner-layer MDP. In each sub-state transition, the sub-action of the agent is to either select one o-d pair, or to issue a holding signal to stop dispatching in the current batch. When all pairs are assigned or the holding signal is issued, the inner-layer transition of this batch is finished. Then the outer-layer MDP enters next state with new drivers and passengers in the upcoming batch. This process repeats till the end of the micro-view time period. Naturally, this cascaded two-layer architecture transforms a series of CO tasks across batches into a sequential decision-making problem that can be solved by DRL. We further describe the two-layer MDP in Sec. \ref{sec21}, Sec. \ref{sec22} and present notations in Table \ref{tab:notation} for clarity.

\subsubsection{Outer-layer MDP}\label{sec21}
The outer-layer MDP is defined as follows:
\begin{itemize}[leftmargin=*]
    \item \textbf{Agent}. The centralized order-dispatching system is modeled as an agent and MICOD follows a single-agent setting.
    
    \item \textbf{State} $\mathcal{S}$. The state $s_t \in \mathcal{S}$ at batch $t$ consists of global-level and local-level information. Formally, $s_t = (I_g^t, I_p^t)$, where $I_g^t$ is the real-time D\&S information in the geo-fence, and $I_p^t$ represents all available o-d pairs with contextual attributes (spatiotemporal distributions, behavioral patterns). Note that $I_p^t$ is of unfixed size since the number of o-d pairs is unspecified in each batch.
    
    \item \textbf{Action} $\mathcal{A}$. The action $a_t$ represents the combinatorial o-d assignment decision in each batch $t$, i.e., a set of o-d pairs from $I_p^t$. Let $n$ be the total number of assigned pairs in a batch. Note that $n$ is unspecified and hardly the same in different batches (elaborated in Sec. \ref{sec22}). In fact, not all available drivers or orders should be assigned exhaustively in each batch as a global optimal solution. To achieve this, a "hold" action similar to StH will be described in Sec. \ref{sec22} as our solution.

    \item \textbf{Policy}. The policy $\pi(a_t|s_t)$ specifies the probability of selecting a combination of pairs as action $a_t$ given the state $s_t$, and is a composition of sub-policies described in Sec. \ref{sec22}.
    
    \item \textbf{State transition} $\mathcal{P}$, \textbf{Initial state distribution} $\rho_0$. The state transition function $\mathcal{P}: \mathcal{S} \times \mathcal{A} \rightarrow \mathcal{S}$ captures the results and bilateral uncertainty induced by the workflow in the outer-layer. The initial state distribution $\rho_0$ describes the spatiotemporal and contextual distribution in each MICOD environment.

    \item \textbf{Reward}. The reward $r_t$ is aligned with different optimization goals: $r_t$ can be defined as the total prices of assigned orders when maximizing driver income or negative pickup distances of assigned o-d pairs when optimizing passenger experience. Note that $r_t$ is received only when a complete action $a_t$ is finished.
\end{itemize}

The objective of the outer-layer MDP is to maximize the discounted return, given policy $\pi$, in the MICOD context within all T steps (the number of all batches). We can formulate the objective with discount factor $\gamma$: 
\begin{equation}\label{eq211}
\setlength{\belowdisplayskip}{-3mm}
\textit{\rm{max}} \; \mathbb{E}_{\pi}[\sum_{t = 0}^T \gamma^t  r_t]
\end{equation}
\begin{table}[t]
\begin{center}
  \setlength{\abovecaptionskip}{0mm}
  \caption{Two-layer MDP Notations}
  \label{tab:notation}
  \begin{adjustbox}{width=0.96\columnwidth}
  \begin{tabular}{cl}
    \toprule
    Notation & Description\\
    \midrule
    $t$ & a batch, i.e., a decision time window \\
    
    $i$ & sub-step in each inner-layer MDP \\
    
    $s$ & state in the outer-layer, $s = (I_g, I_p)$\\
    
    $I_g$ & global contextual information \\
    
    $I_p$ & available o-d pairs information \\
    
    $u$ & sub-state in the inner-layer \\
    
    $a$ & matching decisions in the outer-layer \\
    
    $a^i$ & sub-action in the inner-layer, $a^i=(h^i, c^i)$ \\
    
    $c^i$ & a sub-action category: select an o-d pair \\
    
    $h^i$ & a sub-action category: end matching or not \\
  \bottomrule
\end{tabular}
\end{adjustbox}
\end{center}
\vspace{-4mm}
\end{table}
\subsubsection{Inner-layer MDP}\label{sec22}
The inner-layer MDP models a CO problem in each batch, where unspecified number of drivers and orders are presented. We decompose this task into a sequence of sub-actions that iteratively select a proper o-d pair, remove related pairs (thus naturally conform with one-to-one restriction) and enter a new sub-state, until a certain ending criterion is met. The inner-layer MDP respects the outer-layer MDP setting and highlights the following complementary specifics:
\begin{itemize}[leftmargin=*]
    \item \textbf{Agent} The inner-layer MDP agent is exactly the centralized order-dispatching system in the outer-layer MDP.
    
    \item \textbf{Sub Action}. We define two categories of sub-actions. First, the agent chooses an o-d pair from the pool of available pairs, dubbed as the \textbf{D}ecision. In this way, drivers and orders associated with the chosen pair will be removed from the pool. Second, the agent predicts whether to end the current dispatching process or not, dubbed as the \textbf{H}old. We denote them as $c$ and $h$, respectively. Thus, a sub-action is two-dimensional and $a^i = (h^i, c^i)$ at sub-step $i$ in a batch $t$. Let $n_i$ be the number of remaining o-d pairs at sub-step $i$, then $c^i \in [1, \cdots, n_i]$. $h^i \in \{0,1\}$ and is a binary sub-action, where 1 stands for "hold". 

    In the inner-layer MDP, the agent repeats this sub-action $a_i$: it determines whether to end the dispatching process with a Hold sub-action ($h_i$), then chooses a pair with a Decision sub-action ($c_i$) if the process is not ended, removes the associated pairs from the pool. The two modules are in parallel, and once the Hold module waves an ending signal, the current batch is ceased. 
    
    Thus, a complete action $a$ in the outer-layer MDP is a combination of dependent sequential sub-actions $a_i$. The action space of $a$ is factored \cite{metz2017discrete} and expressed as a Cartesian product of sub-action spaces: $\mathcal{A} = \mathcal{A}_1 \times \cdots \times \mathcal{A}_n$, where $n$ is the total number of sub-steps in a batch, and each $\mathcal{A}_i$ is a two-dimensional discrete space of size $|A_i| = n_i + 2$ (hold sub-action is binary). 

    \item \textbf{Sub State}. The sub-states are closely related to sub-actions. At sub-step $i$, the sub-state space $\mathcal{U}_i = \mathcal{S} \times \mathcal{A}_1 \times \cdots \mathcal{A}_i$, and $\mathcal{U}_0 = \mathcal{S}$. A sub-state $u^i \in \mathcal{U}_i$ represents intermediate dynamics of state $s \in \mathcal{S}$ and all previously generated sub-actions $a^1, \cdots, a^{i-1}$. It encompasses the information of chosen and remaining o-d pairs along with current D\&S contexts, providing decision-making support for the agent.

    \item \textbf{Inner-layer Reward}. The inner-layer reward is set to 0 within each sub-state transition, and the same reward $r_t$ as in the outer-layer is received when a complete action $a_t$ is finished.
    
    \item \textbf{Sub Policy}. Policy $\pi$: $\mathcal{S} \rightarrow \mathcal{A}$ is further decomposed into $n$ corresponding sub-policies $\pi^i: \mathcal{U}_{i-1} \rightarrow \mathcal{A}_i, i \in [1, \cdots, n]$ ($n$ is unfixed). Formally, the probability of selecting a sub-action $a^i$ is expressed as $\pi^i(a^i|s, a^1, \cdots, a^{i-1})$, and that of a complete action $a_t$ is $\pi(a_t|s_t) = \prod_{i=1}^n\pi^i(a_t^i|u_t^{i-1})$ at batch $t$. 
    
    Since each sub-action is two-dimensional, two types of sub-state transitions are traced by sub-policies accordingly. First, $\pi^i$ (recall the Decision) iteratively chooses o-d pairs from the pool unless the Hold module indicates to end. Second, $\pi^i$ (recall the Hold) decides to early stop the current dispatching process and enter the next batch. Thus at sub-step $i$ of batch $t$, we have 
\end{itemize}
\begin{equation}\label{eq221}
\pi^i(a_t^i|u_t^{i-1}) = \left\{
        \begin{array}{ll}
            \pi^i(h_t^i|u_t^{i-1}) \cdot \pi^i(c_t^i|u_t^{i-1}) & i < n \\
            \pi^i(h_t^i|u_t^{i-1}) & i = n
        \end{array}
    \right.
\end{equation}
\subsection{Deep Double Scalable Network}\label{sec:method}
There are two levels of action space scalability in the setting of the two-layer MDP. First, the outer-layer action is scalable since the total number $n$ of assigned o-d pairs is unidentified in advance for each batch. Second, the inner-layer sub-actions are scalable because at each sub-step $i$, the sub-action space size $n_i + 2$ decreases according to previous assignments.

\begin{figure}[t!]
    \setlength{\belowcaptionskip}{-5mm}
    \centering
    \includegraphics[width=\linewidth]{{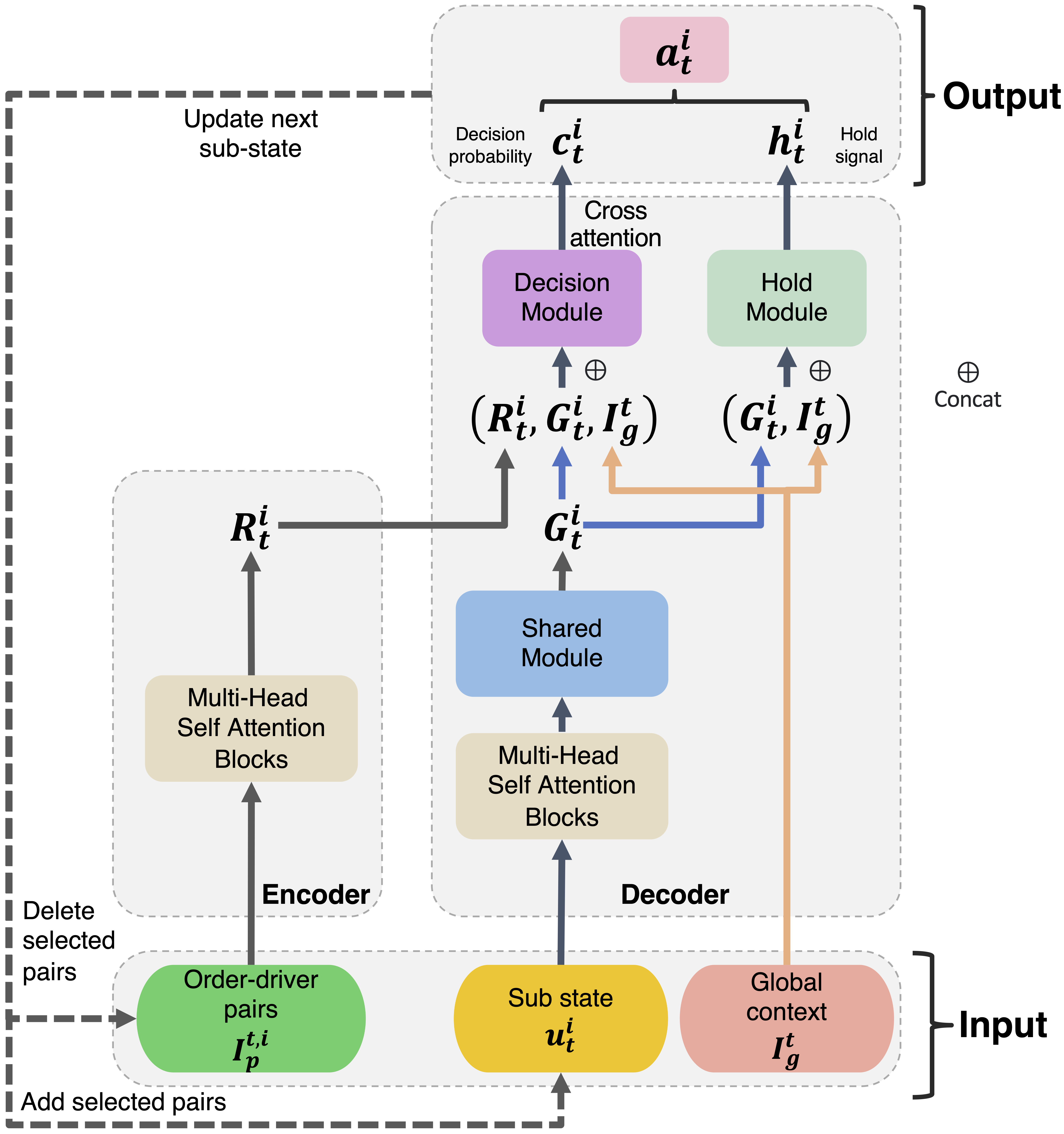}}
    \caption{The architecture of D2SN. The state is updated after each sub-step $i$ of batch $t$ auto-regressively.}\label{model}
\end{figure}

Inspired by sequence-generating tasks in Natural Language Processing, we regard each o-d pair as a "word" for generation. Such analogy is based on the fact that each o-d pair assignment is conditioned on previous assignments, and it will also influence future assignments. Besides, the generation of a sensible sentence relies on each word, just as an overall optimization is based on each o-d assignment. Therefore, we adopt an encoder-decoder structure to generate assignments, with auto-regressive factorization \cite{pierrot2020factored} to adapt to the first level of scalability. The multiple attention mechanisms enable the model to adapt to the second scalability on sub-actions. The proposed architecture is named as \underline{D}eep \underline{D}ouble \underline{S}calable \underline{N}etwork (D2SN), as shown in Figure \ref{model}.

\subsubsection{Encoder}
At each sub-step $i$ of batch $t$, the encoder $E$ takes the input of contextual information of all o-d pairs, i.e., $I_p^{t,i}$, and outputs a latent embedding $R_t^i$ of all pairs. For $n_i$ o-d pairs at sub-step $i$ with feature dimension $d$ for each pair, we have
\begin{equation}\label{eq3111}
\begin{aligned}
R_t^i = E(I_p^{t,i}) = FFN(Attention(I_p^{t,i})),  \;\; R_t^i \in \mathbb{R}^{n_i\times d}
\end{aligned}
\end{equation}
$Attention(\cdot)$ and $FFN(\cdot)$ represent multi-head attention (MHA) blocks and position-wise feed-forward operations, respectively. With such aids, the size of the output $R_t^i$ keeps consistent with that of the input $I_p^{t,i}$, which is of scalable size.

\subsubsection{Decoder}\label{secdecoder}
The decoder takes $R_t^i$ and the global contextual information $I_g^t$ as the input, and outputs (i) the decision probability of each o-d pair and (ii) the hold action decision.

In the decoder, in addition to the MHA blocks, we design a shared module to normalize feature dimensions before the two action modules. As depicted in Equation (\ref{eq3121}), a recurrent block $RNN$ converts the unfixed size of sub-state representation $u_t^i$ to a fixed size vector $G_t^i$, preserving the chronological information of sub-actions. Note that at each sub-step $i$, the decoder input $u_t^i$ is a concatenation of the initial sub-state $u_t^0 = s_t$ (of size $n_0$) and selected pairs $(c_t^1, \cdots, c_t^{i-1})$:
\begin{equation}\label{eq3121}
   \begin{aligned}
   G_t^i &= RNN(Attention(u_t^i)), \; G_t^i \in \mathbb{R}^{1 \times d}, u_t^i \in \mathbb{R}^{(n_0+i-1) \times d} \\
   \end{aligned}
\end{equation}

Then, $G_t^i$ is leveraged by two parallel sub-action modules:
\begin{itemize}[leftmargin=*]
    \item Hold module $H$ takes $G_t^i$ and $I_g^t$ as input, then outputs a binary distribution $h_t^i$ to decide whether to end the current batch:
    \begin{equation}
   \begin{aligned}
   h_t^i &= H(G_t^i, I_g^t), \; h_t^i \in \mathbb{R}^2 \\
   \end{aligned}
   \end{equation}
    \item Decision module $C$ receives the triplet of $(G_t^i$, $I_g^t$, $R_t^i)$. We apply a cross-attention operation between the fixed-size decoder query ($G_t^i, I_g^t \in \mathbb{R}^{1 \times d}$) and the scalable-size encoder keys ($R_t^i \in \mathbb{R}^{n_i \times d}$) to decide which pair is selected from the $n_i$ pairs in $I_p^{t,i}$:
    \begin{equation}
        \begin{aligned}
        c_t^i &= C(R_t^i, G_t^i, I_g^t), \; c_t^i \in \mathbb{R}^{n_i}\\
        \end{aligned}
   \end{equation}
\end{itemize}

\subsubsection{Cooperation}
We enable the auto-regressive sub-action sampling within the encoder-decoder cooperation. Different components of $\{I_p^{t,i}, u_t^i, I_g^t\}$ in state $s_t$ are passed into the encoder and the decoder separately.
D2SN generates $(h_t^i, c_t^i)$ of two parallel sub-actions at each sub-step $i$. If $h_t^i$ is not to hold, then in the inner-layer (i) $I_p^{t,i}$ is renewed by removing pairs related to the selected o-d pair of $c_t^i$, and (ii) $u_t^i$ is updated to $u_t^{i+1}$ by concatenating $\{c_t^1, \cdots, c_t^{i-1}\}$ and $u_t^0$. This process is repeated until no available pairs in $I_p^{t,i}$, or $h_t^i$ indicates to stop.

\subsection{Deep Reinforcement Learning with D2SN}\label{sec32}
Our two-layer MDP adopts clipped proximal policy optimization (PPO) \cite{schulman2017proximal}. DRL operates in the outer-layer, while the inner-layer is modeled with the proposed D2SN architecture.

\subsubsection{Actor-Critic Design}
The training consists of two main networks: an auto-regressive actor (D2SN) and a value-independent critic. The critic utilizes the same decoder structure in D2SN as in Sec. \ref{secdecoder}, but without the two action modules. It only receives $s_t=(I_p^t, I_g^t)$ as the input. Sub-states and actions are not fed into the critic to ensure value-independence.

\subsubsection{DRL Training}
Similar to clipped PPO, we use a surrogate objective loss:
\begin{equation}\label{eq421}
L_{clip}(\theta) = \hat{\mathbb{E}}_t[\textit{\rm{min}}(pr_t(\theta)\hat{A}_t, \textit{\rm{clip}}(pr_t(\theta), 1 - \epsilon, 1+ \epsilon)\hat{A}_t)],
\end{equation}
where $\hat{\mathbb{E}}_t$ indicates the empirical average over a finite batch of samples. $pr_t(\theta) = \frac{\pi_{\theta}(a_t|s_t)}{\pi_{\theta_{old}}(a_t|s_t)}$ denotes the probability ratio, and the clip function truncates $pr_t(\theta)$ to the range of ($1 - \epsilon, 1 + \epsilon$). $\hat{A}_t$ is the advantage estimator at batch $t$. Equation (\ref{eq221}) is plugged into $pr_t(\theta)$ for exact transition probability calculation. Since all sub-actions are sampled auto-regressively, the resulting $\pi_{\theta}(a_t|s_t)$ and $\pi_{\theta_{old}}(a_t|s_t)$ are computed as the product given by the sub-distributions over the action components in sequence:
\begin{equation}\label{eq3221}
\begin{aligned}
    \pi_{\theta}(a_t|s_t) = \Pi_{i=0}^{n}\pi^i(a_t^i|u_t^{i-1}),
\end{aligned}
\end{equation}

Network parameters $\theta$ of the actor are updated by maximizing Equation (\ref{eq421}). Besides, the network parameters $\phi$ of the critic $V$ is trained with Generalized Advantage Estimation (GAE) \cite{schulman2017proximal}:
\begin{equation}\label{eq422}
    \begin{aligned}
    \hat{A}_t &= \delta_t + (\gamma\lambda)\delta_{t+1} + \cdots + (\gamma\lambda)^{T-t+1}\delta_{T-1},\\
    \delta_t &= r_t + \gamma V(s_{t+1}) - V(s_t), \\
    L_c(\phi) &= [V_{\phi}(s_t) - (\hat{A}_t + V_{\phi_{old}}(s_{t+1})),
\end{aligned}
\end{equation}
where $\lambda$ is a hyper-parameter to balance variance and bias and $\phi_{old}$ denotes parameters before updating in each iteration.

We describe the training process in Algorithm \ref{alg1}. Note that network parameters $\phi, \theta$ are optimized with mini-batch $Adam$ \cite{choi2019empirical}).

\begin{algorithm}
\caption{D2SN with two-layer MDP framework for MICOD}\label{alg1}
\begin{algorithmic}[1]
\State Initialize network parameters $\phi, \theta$
\For{\textit{iteration} $k = 1, \cdots, K$} 
    \State Set $\theta_{old} \gets \theta$, reset $\mathcal{D}$
    \For{\!\textbf{each} \textit{batch} $t = 1, 2, \cdots, T$}
        \For{\textit{sub-step} $i = 1, 2, \cdots$}
            \If{not terminating}
                \State Sample a sub-action $(h_t^i, c_t^i)$ using $\pi_{\theta_{old}}$,
                \State and update $I_p^t$ to observe the sub-state $u_t^{i+1}$
                \State Collect instance $(u_t^i, h_t^i, c_t^i, u_t^{i+1})$
            \Else{
                Concatenate all $n$ instances} 
            \EndIf
        \EndFor
    \State Execute $[c_t^i]_{i=1}^n \in a_t$ and receive $r_t, s_{t+1}$
    \State Store sample $(s_t, a_t, s_{t+1}, r_t)$ in $\mathcal{D}$
    \EndFor
    \State Compute advantages $\hat{A}_1, \cdots, \hat{A}_T$ by Equation (\ref{eq422})
    \State Compute sample gradient of $L_c(\phi)$ in Equation (\ref{eq422})
    \State Compute sample gradient of $L_{clip}(\theta)$ in Equation (\ref{eq421})
    \State Update actor $\theta \gets Adam(\theta, \nabla L_{clip}(\theta))$
    \State Update critic $\phi \gets Adam(\phi, \nabla L_c(\phi))$
\EndFor
\end{algorithmic}
\end{algorithm}
\vspace{-3mm}
\section{Experiments}\label{sec:experiments}
\subsection{Experiment Setting}\label{sec51}
\subsubsection{Industrial Benchmarks}
We rewind real-world trajectories in multiple cities of China in 2023 from Didi, then split them into 10-minute dataset each. The benchmarks are classified into 12 unique types based on the combination of D\&S ratios (order numbers/driver numbers) and ranges of driver capacity (e.g., "$\leq400$" means samples with no more than 400 drivers in 10 minutes) as in Table \ref{benchmark}. Overall, benchmarks consist of 875 for training, 221 for validation, 12 for test and encompass over 30 features, including drivers' idle movement records, orders' trip duration, etc. They are formally treated as standard datasets in Didi for offline evaluation.

\begin{table}[!t]
    \setlength{\abovecaptionskip}{1mm}
    \caption{D\&S ratios and driver number ranges of benchmarks}
    \label{benchmark}
    \centering
    \begin{adjustbox}{width=0.94\columnwidth}
    \begin{tabular}{ccccccccccccc}
    \hline
    \multicolumn{1}{|c|}{\textbf{Level}} & \multicolumn{3}{c|}{\textbf{L1}} & \multicolumn{3}{c|}{\textbf{L2}} & \multicolumn{3}{c|}{\textbf{L3}} & \multicolumn{3}{c|}{\textbf{L4}}\\
    \multicolumn{1}{|c|}{D\&S ratios} & \multicolumn{3}{c|}{1.0-1.1} & \multicolumn{3}{c|}{1.1-1.5} & \multicolumn{3}{c|}{1.5-2.0} & \multicolumn{3}{c|}{2.0-4.0}\\
    \hline
    \multicolumn{1}{|c|}{\textbf{Driver Capacity}} & \multicolumn{4}{c|}{$\mathbf{\leq400}$} & \multicolumn{4}{c|}{$\mathbf{\leq550}$} & \multicolumn{4}{c|}{$\mathbf{\leq800}$} \\
    \multicolumn{1}{|c|}{Range} & \multicolumn{4}{c|}{300-400} & \multicolumn{4}{c|}{400-550} & \multicolumn{4}{c|}{550-800} \\
    \hline
    \end{tabular}
    \end{adjustbox}
    \vspace{-4mm}
\end{table}

\subsubsection{Simulator}
An MICOD-customized simulator from Didi operates in a batch mode of 2 seconds for each 10-minute dataset:
\begin{itemize}[leftmargin=*]
    \item It generates orders and drivers from the dataset every 2 seconds and follows the pipeline of "Filtering + Matching + Serving".
    \item It simulates contextual dynamics and behavioral patterns in each batch. With careful calibration, the differences of multiple funnel metrics (answer, complete) between the "real-world" and "simulation" for L1-L4 are all within 1.5\%.
\end{itemize}

Note that the simulator serves as an official and trustworthy indicator in Didi for online policy evaluation as it supports multiple deployed policies, including several versions of StH.

\subsubsection{Baselines}
We take 6 competitive baselines. Naive CO methods contains the Greedy \cite{kalyanasundaram1993online}, Kuhn-Munkres (KM) \cite{kuhn1955hungarian} and Gale-Sharpley (GS) \cite{gale1962college}, they are fully used in Didi's scenarios. Two-stage approaches includes Restricted Q-learning (RQL) \cite{wang2019adaptive}, Interval Delay (ID) \cite{qin2021optimizing} and StH (widely deployed in Didi):
\begin{itemize}[leftmargin=*]
    \item \textbf{RQL} is two-stage. We apply it with DRL framework and learn a batch-splitting policy to control the matching frequency.
    
    \item \textbf{ID} learns delay-matching policy with DRL framework. ID differs from RQL in that ID partitions a geo-fence into pieces in advance based on the density of requests.
    
    \item \textbf{StH} follows the intuition of optimal stopping by predicting best matching moments for drivers or orders in a DL manner.
\end{itemize}

\subsubsection{Evaluation Metrics} We focus on two different tasks:

\noindent \textbf{Average Pickup Distance (APD).} In this passenger-view task:
\begin{itemize}[leftmargin=*]
    \item Reward $r_t$ is defined as the total negative pickup distances of served orders in each batch $t$.

    \item The objective is to maximize the complete ratio (CR, i.e., finished orders out of all appeared orders) and minimize the average pickup distance of all served orders.
\end{itemize}

\noindent \textbf{Total Driver Income (TDI).} In this driver-view task:
\begin{itemize}[leftmargin=*]
    \item Reward $r_t$ is the total prices of served orders in each batch $t$.

    \item The objective considers both the CR and the income of all drivers.
\end{itemize}
The APD task is tested on L1 samples of balanced D\&S ratios as they stand for off-peak periods in which passenger experience is emphasized. The TDI task is evaluated across L2-L4 samples of unbalanced D\&S ratios as they represent various peak periods in which CR and driver experience are together stressed.

\subsubsection{Implementation Details} DRL baselines (RQL, ID, D2SN) are optimized with the same training strategy and input features. They all converge within a maximum of 50 epochs, and we use the best model of each method for test. The o-d feature embedding dimension of all models in this paper are set to 256 and the learning rate is set to 0.002 with $Adam$ \cite{choi2019empirical}. Specifically, D2SN contains about 1.5 million parameters and its average execution duration is at the millisecond-level, comparable to other two-stage methods. 

For fair comparison, evaluations of all methods in each test sample are conducted with different seeds over 30 runs. Each metric result in Table \ref{apd}, \ref{gmvcr} is in the form of mean and standard deviation. All experiments are conducted on a server (Ubuntu 18.04) with an Intel Xeon Platinum 8352Y CPU and an Nvidia RTX A6000 GPU.

\subsection{Main Results}\label{sec52}
Table \ref{apd}, \ref{gmvcr} present performances of all methods in terms of two tasks APD and TDI, respectively. Overall, D2SN displays notable advantages over baselines in both tasks. Specifically, D2SN outperforms the deployed policy StH in all L1-L4 scenarios, indicating its trustworthy potential of online performance.

In the APD task, D2SN slightly lags behind RQL and ID in reducing pickup distance while keeps the highest CR in all L1 tasks. That means D2SN is capable of jointly satisfying order requests (CR) and pickup experience over the entire time period rather than merely focuses on downsizing pickup distances as ID and RQL do. 

For the TDI task in Table \ref{gmvcr}, D2SN gains an robust improvement of $0.7\%-3.90\%$ and around $1\%-2\%$ over the best of other baselines in terms of TDI and CR, respectively. In particular, as the D\&S ratios grow higher, i.e., from L2 to L4, the improvement of D2SN over other methods increases proportionally. This is because with higher D\&S ratios, D2SN is provided with more decision space for o-d level manipulation. The advantage in CR again demonstrates that D2SN is able to capture contextual changes as the model implicitly predicts cancellation behaviors of drivers and orders.
\begin{table}[!t]
    \setlength{\abovecaptionskip}{1mm}
    \caption{APD task performance. In each entry, the upper is CR and the lower is APD. In each column, bold values denote the highest CR and the lowest (best) APD.}
    \label{apd}
    \centering
    \begin{tabular}{cccc}
    \toprule
    \multicolumn{1}{c}{\multirow{2}{*}{Benchmark}} & \multicolumn{3}{c}{L1 (1.0-1.1)} \\ [0.05ex]
    \cmidrule(rl){2-4}
    \multicolumn{1}{c}{} & \multicolumn{1}{c}{$\leq400$} & \multicolumn{1}{c}{$\leq550$} & \multicolumn{1}{c}{$\leq800$} \\ [0.05ex]
    \midrule
    \multicolumn{1}{c}{\multirow{2}{*}{Greedy}} & $0.80\pm0.01$ &  $0.81\pm0.01$ &  $0.83\pm0.01$ \\ [0.05ex]
    \multicolumn{1}{c}{}  & $1493\pm53$ & $1683\pm35$ & $1604\pm36$ \\ [0.05ex] 
    \hline
    \multicolumn{1}{c}{\multirow{2}{*}{KM}} & $0.80\pm0.01$ & $0.80\pm0.01$ & $0.83\pm0.008$ \\ [0.05ex]
    \multicolumn{1}{c}{} &  $1383\pm33$ & $1504\pm25$ & $1319\pm24$\\ [0.05ex]
    \hline
    \multicolumn{1}{c}{\multirow{2}{*}{GS}} & $0.80\pm0.01$ & $0.80\pm0.01$ & $0.82\pm0.006$  \\ [0.05ex]
    \multicolumn{1}{c}{} & $1463\pm48$ & $1614\pm35$ & $1527\pm24$\\ [0.05ex]
    \hline
    \multicolumn{1}{c}{\multirow{2}{*}{RQL\cite{wang2019adaptive}}} & $0.78\pm0.01$ & $0.79\pm0.01$ & $0.80\pm0.01$ \\ [0.05ex]
    \multicolumn{1}{c}{} & \ensuremath{\mathbf{1246\pm32}} & $1338\pm27$ & \ensuremath{\mathbf{1158\pm27}}  \\ [0.05ex]
    \hline
    \multicolumn{1}{c}{\multirow{2}{*}{ID\cite{qin2021optimizing}}} & $0.78\pm0.01$ & $0.78\pm0.01$ & $0.77\pm0.01$ \\ [0.05ex]
    \multicolumn{1}{c}{} & $1310\pm27$ & \ensuremath{\mathbf{1337\pm24}} & $1158\pm28$  \\ [0.05ex]
    \hline
    \multicolumn{1}{c}{\multirow{2}{*}{StH}} & $0.79\pm0.02$ & $0.78\pm0.01$ & $0.77\pm0.02$ \\ [0.05ex]
    \multicolumn{1}{c}{} &  $1284\pm44$ & $1393\pm32$ & $1263\pm37$\\ [0.05ex]
    \hline
    \multicolumn{1}{c}{\multirow{2}{*}{D2SN(ours)}} &  \ensuremath{\mathbf{0.84\pm0.01}} & \ensuremath{\mathbf{0.87\pm0.01}} & \ensuremath{\mathbf{0.85\pm0.02}}\\ [0.05ex]
    \multicolumn{1}{c}{} & $1306\pm50$ & $1429\pm38$ & $1162\pm33$\\ [0.05ex]
    \hline
    \multicolumn{1}{c}{\multirow{2}{*}{D2SN\textsubscript{h\textsuperscript{-}}(ours)}} & $0.80\pm0.02$ & $0.79\pm0.01$ & $0.81\pm0.02$ \\ [0.05ex]
    \multicolumn{1}{c}{} &  $1390\pm39$ & $1498\pm34$ & $1322\pm27$\\ [0.05ex]
    \bottomrule
    \end{tabular}
    \vspace{-4mm}
\end{table}

\subsection{Performance Analysis}
\subsubsection{Ablation Study}\label{sechm}
We conduct ablation study to validate the effectiveness of D2SN by removing the Hold module as a baseline, dubbed as D2SN\textsubscript{h\textsuperscript{-}}. This method dispatches all o-d pairs exhaustively in each batch without holding and is trained in the same way as D2SN does. Results in the last row of Table \ref{apd}, \ref{gmvcr} shows D2SN\textsubscript{h\textsuperscript{-}} performs similar to KM in L1-L4 scenarios, indicating the dispatching ability of the Decision module alone. However, D2SN\textsubscript{h\textsuperscript{-}} lags behind D2SN by $5\%-13\%$ in terms of APD and $1.5\%-5\%$ in terms of TDI, respectively. Such gaps display that D2SN\textsubscript{h\textsuperscript{-}} is unable to reasonably hold for future matching.
\begin{table*}[t]
    \setlength{\abovecaptionskip}{1mm}
    \caption{TDI task performance. In each entry, the upper is TDI and the lower is CR. In each column, bold values denote the highest TDI and CR.}
    \label{gmvcr}
    \centering
    \begin{adjustbox}{width=\textwidth}
    \begin{tabular}{cccccccccc}
    \toprule
    \multicolumn{1}{c}{\multirow{2}{*}{Benchmark}} & \multicolumn{3}{c}{L2 (1.1-1.5)} & \multicolumn{3}{c}{L3 (1.5-2.0)} & \multicolumn{3}{c}{L4 (2.0-4.0)}\\ [0.6ex]
    \cmidrule(rl){2-10}
    \multicolumn{1}{c}{} & \multicolumn{1}{c}{$\leq400$} & \multicolumn{1}{c}{$\leq550$} & \multicolumn{1}{c}{$\leq800$} & \multicolumn{1}{c}{$\leq400$} & \multicolumn{1}{c}{$\leq550$} & \multicolumn{1}{c}{$\leq800$} & \multicolumn{1}{c}{$\leq400$} & \multicolumn{1}{c}{$\leq550$} & \multicolumn{1}{c}{$\leq800$}\\ [0.6ex]
    \midrule
    \multicolumn{1}{c}{\multirow{2}{*}{Greedy}} & $8370\pm145$ & $12159\pm237$ & $14090\pm204$ & $10116\pm299$ &  $16301\pm260$ &  $18317\pm229$ & $9172\pm134$ &  $14259\pm225$ &  $17549\pm228$\\ [0.6ex]
    \multicolumn{1}{c}{} & $0.76\pm0.007$ &  $0.73\pm0.006$ &  $0.80\pm0.01$  & \ensuremath{\mathbf{0.58\pm0.002}} &  $0.59\pm0.002$ &  $0.57\pm0.002$ & $0.41\pm0.003$ &  $0.44\pm0.001$ &  $0.46\pm0.001$\\ [0.6ex]
    \hline
    \multicolumn{1}{c}{\multirow{2}{*}{KM}} & $8298\pm140$ & $12079\pm297$ & $14033\pm279$ & $10134\pm239$ &  $16341\pm228$ &  $18375\pm187$ & $9146\pm146$ &  $14275\pm182$ &  $17525\pm188$\\ [0.6ex]
    \multicolumn{1}{c}{} & $0.76\pm0.007$ &  $0.72\pm0.007$ &  $0.80\pm0.01$  & $0.57\pm0.003$ &  $0.58\pm0.002$ &  $0.57\pm0.001$ & $0.40\pm0.005$ &  $0.44\pm0.001$ &  $0.46\pm0.001$\\ [0.6ex]
    \hline
    \multicolumn{1}{c}{\multirow{2}{*}{GS}} & $8386\pm171$ & $12336\pm317$ & $14336\pm240$ & $10165\pm233$ &  $16263\pm238$ &  $18312\pm206$ & $9172\pm122$ &  $14267\pm196$ &  $17508\pm184$\\ [0.6ex]
    \multicolumn{1}{c}{} & $0.76\pm0.007$ &  $0.73\pm0.007$ &  \ensuremath{\mathbf{0.81\pm0.008}}  & $0.58\pm0.003$ &  $0.58\pm0.003$ &  $0.57\pm0.001$ & $0.40\pm0.003$ &  $0.44\pm0.001$ &  $0.46\pm0.001$\\ [0.6ex]
    \hline
    \multicolumn{1}{c}{\multirow{2}{*}{RQL\cite{wang2019adaptive}}} & $8331\pm151$ & $12122\pm361$ & $13934\pm310$ & $10124\pm240$ &  $16323\pm300$ &  $18345\pm302$ & $9170\pm190$ &  $14297\pm261$ &  $17531\pm296$\\ [0.6ex]
    \multicolumn{1}{c}{} & $0.76\pm0.006$ &  $0.72\pm0.007$ &  $0.80\pm0.01$  & $0.57\pm0.003$ &  $0.58\pm0.002$ &  $0.57\pm0.001$ & $0.40\pm0.003$ &  $0.43\pm0.001$ &  $0.45\pm0.001$\\ [0.6ex]
    \hline
    \multicolumn{1}{c}{\multirow{2}{*}{ID\cite{qin2021optimizing}}} & $8312\pm171$ & $12055\pm292$ & $13917\pm303$ & $10105\pm195$ &  $16233\pm281$ &  $18495\pm189$ & $9175\pm132$ &  $14268\pm211$ &  $17060\pm233$\\ [0.6ex]
    \multicolumn{1}{c}{} & $0.76\pm0.007$ &  $0.72\pm0.007$ &  $0.80\pm0.009$  & $0.57\pm0.003$ &  $0.58\pm0.004$ &  $0.57\pm0.003$ & $0.40\pm0.003$ &  $0.44\pm0.001$ &  $0.45\pm0.001$\\ [0.6ex]
    \hline
    \multicolumn{1}{c}{\multirow{2}{*}{StH}} & $8303\pm104$ & $11996\pm379$ & $13956\pm299$ & $10095\pm262$ &  $16183\pm229$ &  $18512\pm214$ & $9064\pm157$ &  $14196\pm187$ &  $17615\pm198$ \\ [0.6ex]
    \multicolumn{1}{c}{} & $0.76\pm0.006$ & $0.70\pm0.01$ & $0.80\pm0.01$ & $0.57\pm0.003$ & $0.58\pm0.004$ &  $0.57\pm0.003$ & $0.39\pm0.005$ & \ensuremath{\mathbf{0.44\pm0.001}} &  \ensuremath{\mathbf{0.46\pm0.001}} \\ [0.6ex]
    \hline
    \multicolumn{1}{c}{\multirow{2}{*}{D2SN(ours)}} & \ensuremath{\mathbf{8626\pm133}} & \ensuremath{\mathbf{12691\pm263}} & \ensuremath{\mathbf{14441\pm198}} & \ensuremath{\mathbf{10573\pm240}} &  \ensuremath{\mathbf{16837\pm217}} &  \ensuremath{\mathbf{18683\pm194}} & \ensuremath{\mathbf{9530\pm127}} &  \ensuremath{\mathbf{14552\pm202}} &  \ensuremath{\mathbf{17962\pm209}} \\ [0.6ex]
    \multicolumn{1}{c}{} & \ensuremath{\mathbf{0.78\pm0.007}} & \ensuremath{\mathbf{0.74\pm0.01}} & $0.80\pm0.006$ & $0.57\pm0.004$ & \ensuremath{\mathbf{0.59\pm0.002}} &  \ensuremath{\mathbf{0.58\pm0.002}} & \ensuremath{\mathbf{0.41\pm0.002}} & $0.44\pm0.002$ &  $0.46\pm0.002$ \\ [0.6ex]
    \hline
    \multicolumn{1}{c}{\multirow{2}{*}{D2SN\textsubscript{h\textsuperscript{-}}(ours)}} & $8311\pm126$ & $12066\pm235$ & $14007\pm251$ & $10129\pm253$ &  $16315\pm211$ &  $18402\pm227$ & $9133\pm160$ &  $14269\pm209$ &  $17533\pm195$\\ [0.6ex]
    \multicolumn{1}{c}{} & $0.76\pm0.01$ &  $0.72\pm0.01$ &  $0.80\pm0.01$  & $0.57\pm0.004$ &  $0.58\pm0.002$ &  $0.57\pm0.001$ & $0.41\pm0.003$ &  $0.44\pm0.001$ &  $0.46\pm0.001$\\ [0.6ex]
    \bottomrule
    \end{tabular}
    \end{adjustbox}
    \vspace{-3mm}
\end{table*}
\subsubsection{Hold Module Analysis}\label{sechpa}
The well functioning of the Hold module showcases the advantage of D2SN in the ablation. Thus, we further investigate the 4 learning-based approaches with "hold" strategies and compare the metrics below (the term "distinct" refers to each item is counted once even repeatedly appeared):
\begin{itemize}[leftmargin=*]
    \item Hold-APD ratio: the APD of o-d pairs filtered by "hold" strategy, divided by the APD of finished orders.
    \item Hold-O ratio: the number of distinct orders filtered by "hold" strategy, divided by the total number of orders.
    \item Hold-TDI ratio: the average TDI of o-d pairs filtered by "hold" strategy, divided by the average TDI of finished orders.
    \item Hold-D ratio: the number of distinct drivers filtered by "hold" strategy, divided by the total number of drivers.
    \item SR: the number of distinct served individuals, divided by the total number of them (drivers and orders, respectively).
\end{itemize}

\begin{figure}[!h]
    \setlength{\belowcaptionskip}{-2mm}
    \centering
    \includegraphics[width=\linewidth]{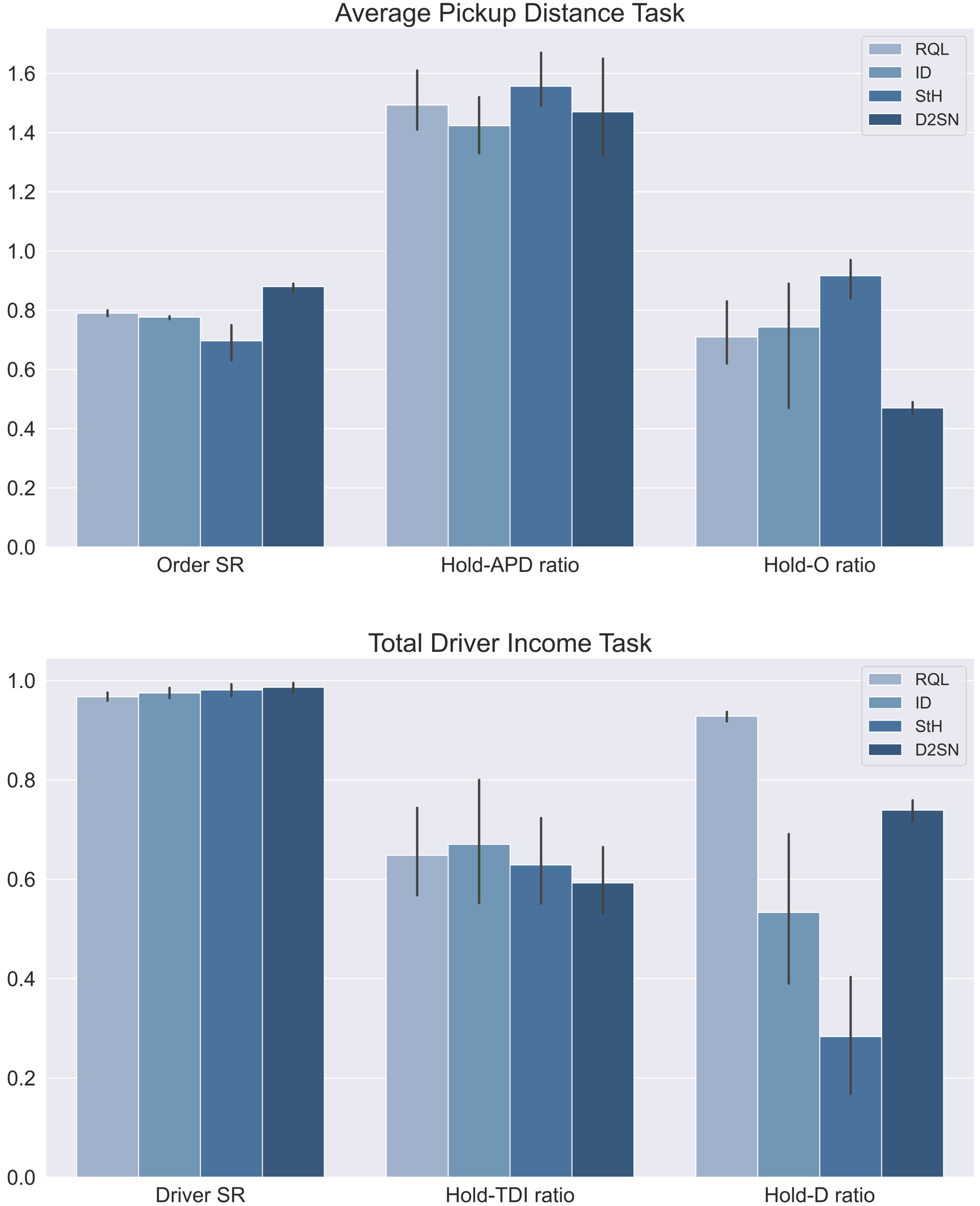}
    \caption{Comparison of methods with "hold" strategies, results are averaged across APD and TDI task test samples, respectively.}\label{holdcombo}
\end{figure}

The upper part of Figure \ref{holdcombo} shows that in all APD tasks, D2SN is competitive among all methods in terms of the Hold-APD ratio, Hold-O ratio and order SR.
The Hold module of D2SN is capable of holding pairs with far pickup distance to downsize APD, providing better passenger experience as the Hold-APD ratios are higher than 1. 
With appropriate Hold-O ratio, the order SR of D2SN is superior to other methods, which means D2SN decreases the risk of order cancellation to satisfy as many order requests as possible.

The lower part of Figure \ref{holdcombo} illustrates all 9 TDI task performance. D2SN performs well in holding pairs that may negatively impact drivers income and possesses the lowest Hold-TDI ratio. Besides, the Hold module is able to delay drivers with appropriate Hold-D ratio and decrease the risk of driver offline behavior. This is because D2SN has the highest driver SR among all methods and performs the best in maximizing TDI as shown in Table \ref{gmvcr}.

\subsection{Deployment Consideration}
D2SN will be applied in typical spatiotemporal contexts within certain geo-fences. For every 10 minutes, we will apply an appropriate D2SN model from 12 base scenarios according to the historical D\&S information of this context. Specifically, in scenarios with extreme dense order requests, parallel-style decoding will be incorporated in the system to downsize the inference latency.
\section{Conclusions}\label{sec:cf}
This study concentrates on MICOD in online ride-hailing, addressing bilateral uncertainty challenges. We propose an end-to-end DRL framework with a novel network D2SN to uniformly resolve prediction, decision-making, and optimization without any CO algorithms. We believe our framework can benefit domains of online resource allocation to solve other practical optimization problems.

\bibliographystyle{ACM-Reference-Format}
\balance
\bibliography{sample-base}


\begin{thebibliography}{32}


\ifx \showCODEN    \undefined \def \showCODEN     #1{\unskip}     \fi
\ifx \showDOI      \undefined \def \showDOI       #1{#1}\fi
\ifx \showISBNx    \undefined \def \showISBNx     #1{\unskip}     \fi
\ifx \showISBNxiii \undefined \def \showISBNxiii  #1{\unskip}     \fi
\ifx \showISSN     \undefined \def \showISSN      #1{\unskip}     \fi
\ifx \showLCCN     \undefined \def \showLCCN      #1{\unskip}     \fi
\ifx \shownote     \undefined \def \shownote      #1{#1}          \fi
\ifx \showarticletitle \undefined \def \showarticletitle #1{#1}   \fi
\ifx \showURL      \undefined \def \showURL       {\relax}        \fi
\providecommand\bibfield[2]{#2}
\providecommand\bibinfo[2]{#2}
\providecommand\natexlab[1]{#1}
\providecommand\showeprint[2][]{arXiv:#2}

\bibitem[Chen et~al\mbox{.}(2018)]%
        {chen2018pixelsnail}
\bibfield{author}{\bibinfo{person}{Xi Chen}, \bibinfo{person}{Nikhil Mishra}, \bibinfo{person}{Mostafa Rohaninejad}, {and} \bibinfo{person}{Pieter Abbeel}.} \bibinfo{year}{2018}\natexlab{}.
\newblock \showarticletitle{Pixelsnail: An improved autoregressive generative model}. In \bibinfo{booktitle}{\emph{International Conference on Machine Learning}}. PMLR, \bibinfo{pages}{864--872}.
\newblock


\bibitem[Choi et~al\mbox{.}(2019)]%
        {choi2019empirical}
\bibfield{author}{\bibinfo{person}{Dami Choi}, \bibinfo{person}{Christopher~J Shallue}, \bibinfo{person}{Zachary Nado}, \bibinfo{person}{Jaehoon Lee}, \bibinfo{person}{Chris~J Maddison}, {and} \bibinfo{person}{George~E Dahl}.} \bibinfo{year}{2019}\natexlab{}.
\newblock \showarticletitle{On empirical comparisons of optimizers for deep learning}.
\newblock \bibinfo{journal}{\emph{arXiv preprint arXiv:1910.05446}} (\bibinfo{year}{2019}).
\newblock


\bibitem[Gale and Shapley(1962)]%
        {gale1962college}
\bibfield{author}{\bibinfo{person}{David Gale} {and} \bibinfo{person}{Lloyd~S Shapley}.} \bibinfo{year}{1962}\natexlab{}.
\newblock \showarticletitle{College admissions and the stability of marriage}.
\newblock \bibinfo{journal}{\emph{The American Mathematical Monthly}} \bibinfo{volume}{69}, \bibinfo{number}{1} (\bibinfo{year}{1962}), \bibinfo{pages}{9--15}.
\newblock


\bibitem[Han et~al\mbox{.}(2022)]%
        {han2022real}
\bibfield{author}{\bibinfo{person}{Benjamin Han}, \bibinfo{person}{Hyungjun Lee}, {and} \bibinfo{person}{S{\'e}bastien Martin}.} \bibinfo{year}{2022}\natexlab{}.
\newblock \showarticletitle{Real-Time Rideshare Driver Supply Values Using Online Reinforcement Learning}. In \bibinfo{booktitle}{\emph{Proceedings of the 28th ACM SIGKDD Conference on Knowledge Discovery and Data Mining}}. \bibinfo{pages}{2968--2976}.
\newblock


\bibitem[Ho and Vaughan(2012)]%
        {ho2012online}
\bibfield{author}{\bibinfo{person}{Chien-Ju Ho} {and} \bibinfo{person}{Jennifer Vaughan}.} \bibinfo{year}{2012}\natexlab{}.
\newblock \showarticletitle{Online task assignment in crowdsourcing markets}. In \bibinfo{booktitle}{\emph{Proceedings of the AAAI conference on artificial intelligence}}, Vol.~\bibinfo{volume}{26}. \bibinfo{pages}{45--51}.
\newblock


\bibitem[Holler et~al\mbox{.}(2019)]%
        {holler2019deep}
\bibfield{author}{\bibinfo{person}{John Holler}, \bibinfo{person}{Risto Vuorio}, \bibinfo{person}{Zhiwei Qin}, \bibinfo{person}{Xiaocheng Tang}, \bibinfo{person}{Yan Jiao}, \bibinfo{person}{Tiancheng Jin}, \bibinfo{person}{Satinder Singh}, \bibinfo{person}{Chenxi Wang}, {and} \bibinfo{person}{Jieping Ye}.} \bibinfo{year}{2019}\natexlab{}.
\newblock \showarticletitle{Deep reinforcement learning for multi-driver vehicle dispatching and repositioning problem}. In \bibinfo{booktitle}{\emph{2019 IEEE International Conference on Data Mining (ICDM)}}. IEEE, \bibinfo{pages}{1090--1095}.
\newblock


\bibitem[Hu and Zhou(2022)]%
        {hu2022dynamic}
\bibfield{author}{\bibinfo{person}{Ming Hu} {and} \bibinfo{person}{Yun Zhou}.} \bibinfo{year}{2022}\natexlab{}.
\newblock \showarticletitle{Dynamic type matching}.
\newblock \bibinfo{journal}{\emph{Manufacturing \& Service Operations Management}} \bibinfo{volume}{24}, \bibinfo{number}{1} (\bibinfo{year}{2022}), \bibinfo{pages}{125--142}.
\newblock


\bibitem[Jin et~al\mbox{.}(2019)]%
        {jin2019coride}
\bibfield{author}{\bibinfo{person}{Jiarui Jin}, \bibinfo{person}{Ming Zhou}, \bibinfo{person}{Weinan Zhang}, \bibinfo{person}{Minne Li}, \bibinfo{person}{Zilong Guo}, \bibinfo{person}{Zhiwei Qin}, \bibinfo{person}{Yan Jiao}, \bibinfo{person}{Xiaocheng Tang}, \bibinfo{person}{Chenxi Wang}, \bibinfo{person}{Jun Wang}, {et~al\mbox{.}}} \bibinfo{year}{2019}\natexlab{}.
\newblock \showarticletitle{Coride: joint order dispatching and fleet management for multi-scale ride-hailing platforms}. In \bibinfo{booktitle}{\emph{Proceedings of the 28th ACM International Conference on Information and Knowledge Management}}. \bibinfo{pages}{1983--1992}.
\newblock


\bibitem[Kalyanasundaram and Pruhs(1993)]%
        {kalyanasundaram1993online}
\bibfield{author}{\bibinfo{person}{Bala Kalyanasundaram} {and} \bibinfo{person}{Kirk Pruhs}.} \bibinfo{year}{1993}\natexlab{}.
\newblock \showarticletitle{Online weighted matching}.
\newblock \bibinfo{journal}{\emph{Journal of Algorithms}} \bibinfo{volume}{14}, \bibinfo{number}{3} (\bibinfo{year}{1993}), \bibinfo{pages}{478--488}.
\newblock


\bibitem[Karp et~al\mbox{.}(1990)]%
        {karp1990optimal}
\bibfield{author}{\bibinfo{person}{Richard~M Karp}, \bibinfo{person}{Umesh~V Vazirani}, {and} \bibinfo{person}{Vijay~V Vazirani}.} \bibinfo{year}{1990}\natexlab{}.
\newblock \showarticletitle{An optimal algorithm for on-line bipartite matching}. In \bibinfo{booktitle}{\emph{Proceedings of the twenty-second annual ACM symposium on Theory of computing}}. \bibinfo{pages}{352--358}.
\newblock


\bibitem[Ke et~al\mbox{.}(2020)]%
        {ke2020learning}
\bibfield{author}{\bibinfo{person}{Jintao Ke}, \bibinfo{person}{Feng Xiao}, \bibinfo{person}{Hai Yang}, {and} \bibinfo{person}{Jieping Ye}.} \bibinfo{year}{2020}\natexlab{}.
\newblock \showarticletitle{Learning to delay in ride-sourcing systems: a multi-agent deep reinforcement learning framework}.
\newblock \bibinfo{journal}{\emph{IEEE Transactions on Knowledge and Data Engineering}} \bibinfo{volume}{34}, \bibinfo{number}{5} (\bibinfo{year}{2020}), \bibinfo{pages}{2280--2292}.
\newblock


\bibitem[Kuhn(1955)]%
        {kuhn1955hungarian}
\bibfield{author}{\bibinfo{person}{Harold~W Kuhn}.} \bibinfo{year}{1955}\natexlab{}.
\newblock \showarticletitle{The Hungarian method for the assignment problem}.
\newblock \bibinfo{journal}{\emph{Naval research logistics quarterly}} \bibinfo{volume}{2}, \bibinfo{number}{1-2} (\bibinfo{year}{1955}), \bibinfo{pages}{83--97}.
\newblock


\bibitem[Li et~al\mbox{.}(2019)]%
        {li2019efficient}
\bibfield{author}{\bibinfo{person}{Minne Li}, \bibinfo{person}{Zhiwei Qin}, \bibinfo{person}{Yan Jiao}, \bibinfo{person}{Yaodong Yang}, \bibinfo{person}{Jun Wang}, \bibinfo{person}{Chenxi Wang}, \bibinfo{person}{Guobin Wu}, {and} \bibinfo{person}{Jieping Ye}.} \bibinfo{year}{2019}\natexlab{}.
\newblock \showarticletitle{Efficient ridesharing order dispatching with mean field multi-agent reinforcement learning}. In \bibinfo{booktitle}{\emph{The world wide web conference}}. \bibinfo{pages}{983--994}.
\newblock


\bibitem[Lin et~al\mbox{.}(2018)]%
        {lin2018efficient}
\bibfield{author}{\bibinfo{person}{Kaixiang Lin}, \bibinfo{person}{Renyu Zhao}, \bibinfo{person}{Zhe Xu}, {and} \bibinfo{person}{Jiayu Zhou}.} \bibinfo{year}{2018}\natexlab{}.
\newblock \showarticletitle{Efficient large-scale fleet management via multi-agent deep reinforcement learning}. In \bibinfo{booktitle}{\emph{Proceedings of the 24th ACM SIGKDD International Conference on Knowledge Discovery \& Data Mining}}. \bibinfo{pages}{1774--1783}.
\newblock


\bibitem[Ma et~al\mbox{.}(2021)]%
        {ma2021hierarchical}
\bibfield{author}{\bibinfo{person}{Yi Ma}, \bibinfo{person}{Xiaotian Hao}, \bibinfo{person}{Jianye Hao}, \bibinfo{person}{Jiawen Lu}, \bibinfo{person}{Xing Liu}, \bibinfo{person}{Tong Xialiang}, \bibinfo{person}{Mingxuan Yuan}, \bibinfo{person}{Zhigang Li}, \bibinfo{person}{Jie Tang}, {and} \bibinfo{person}{Zhaopeng Meng}.} \bibinfo{year}{2021}\natexlab{}.
\newblock \showarticletitle{A hierarchical reinforcement learning based optimization framework for large-scale dynamic pickup and delivery problems}.
\newblock \bibinfo{journal}{\emph{Advances in Neural Information Processing Systems}}  \bibinfo{volume}{34} (\bibinfo{year}{2021}), \bibinfo{pages}{23609--23620}.
\newblock


\bibitem[Metz et~al\mbox{.}(2017)]%
        {metz2017discrete}
\bibfield{author}{\bibinfo{person}{Luke Metz}, \bibinfo{person}{Julian Ibarz}, \bibinfo{person}{Navdeep Jaitly}, {and} \bibinfo{person}{James Davidson}.} \bibinfo{year}{2017}\natexlab{}.
\newblock \showarticletitle{Discrete sequential prediction of continuous actions for deep rl}.
\newblock \bibinfo{journal}{\emph{arXiv preprint arXiv:1705.05035}} (\bibinfo{year}{2017}).
\newblock


\bibitem[PIERROT et~al\mbox{.}(2020)]%
        {pierrot2020factored}
\bibfield{author}{\bibinfo{person}{Thomas PIERROT}, \bibinfo{person}{Valentin Mac{\'e}}, \bibinfo{person}{Jean-Baptiste Sevestre}, \bibinfo{person}{Louis Monier}, \bibinfo{person}{Alexandre Laterre}, \bibinfo{person}{Nicolas Perrin}, \bibinfo{person}{Karim Beguir}, {and} \bibinfo{person}{Olivier Sigaud}.} \bibinfo{year}{2020}\natexlab{}.
\newblock \showarticletitle{Factored Action Spaces in Deep Reinforcement Learning}.
\newblock  (\bibinfo{year}{2020}).
\newblock


\bibitem[Qin et~al\mbox{.}(2021)]%
        {qin2021optimizing}
\bibfield{author}{\bibinfo{person}{Guoyang Qin}, \bibinfo{person}{Qi Luo}, \bibinfo{person}{Yafeng Yin}, \bibinfo{person}{Jian Sun}, {and} \bibinfo{person}{Jieping Ye}.} \bibinfo{year}{2021}\natexlab{}.
\newblock \showarticletitle{Optimizing matching time intervals for ride-hailing services using reinforcement learning}.
\newblock \bibinfo{journal}{\emph{Transportation Research Part C: Emerging Technologies}}  \bibinfo{volume}{129} (\bibinfo{year}{2021}), \bibinfo{pages}{103239}.
\newblock


\bibitem[Qin et~al\mbox{.}(2022)]%
        {qin2022reinforcement}
\bibfield{author}{\bibinfo{person}{Zhiwei~Tony Qin}, \bibinfo{person}{Hongtu Zhu}, {and} \bibinfo{person}{Jieping Ye}.} \bibinfo{year}{2022}\natexlab{}.
\newblock \showarticletitle{Reinforcement learning for ridesharing: An extended survey}.
\newblock \bibinfo{journal}{\emph{Transportation Research Part C: Emerging Technologies}}  \bibinfo{volume}{144} (\bibinfo{year}{2022}), \bibinfo{pages}{103852}.
\newblock


\bibitem[Schulman et~al\mbox{.}(2017)]%
        {schulman2017proximal}
\bibfield{author}{\bibinfo{person}{John Schulman}, \bibinfo{person}{Filip Wolski}, \bibinfo{person}{Prafulla Dhariwal}, \bibinfo{person}{Alec Radford}, {and} \bibinfo{person}{Oleg Klimov}.} \bibinfo{year}{2017}\natexlab{}.
\newblock \showarticletitle{Proximal policy optimization algorithms}.
\newblock \bibinfo{journal}{\emph{arXiv preprint arXiv:1707.06347}} (\bibinfo{year}{2017}).
\newblock


\bibitem[Shiryaev(2007)]%
        {shiryaev2007optimal}
\bibfield{author}{\bibinfo{person}{Albert~N Shiryaev}.} \bibinfo{year}{2007}\natexlab{}.
\newblock \bibinfo{booktitle}{\emph{Optimal stopping rules}}. Vol.~\bibinfo{volume}{8}.
\newblock \bibinfo{publisher}{Springer Science \& Business Media}.
\newblock


\bibitem[Sun et~al\mbox{.}(2022)]%
        {sun2022optimizing}
\bibfield{author}{\bibinfo{person}{Jiahui Sun}, \bibinfo{person}{Haiming Jin}, \bibinfo{person}{Zhaoxing Yang}, \bibinfo{person}{Lu Su}, {and} \bibinfo{person}{Xinbing Wang}.} \bibinfo{year}{2022}\natexlab{}.
\newblock \showarticletitle{Optimizing long-term efficiency and fairness in ride-hailing via joint order dispatching and driver repositioning}. In \bibinfo{booktitle}{\emph{Proceedings of the 28th ACM SIGKDD Conference on Knowledge Discovery and Data Mining}}. \bibinfo{pages}{3950--3960}.
\newblock


\bibitem[Tang et~al\mbox{.}(2019)]%
        {tang2019deep}
\bibfield{author}{\bibinfo{person}{Xiaocheng Tang}, \bibinfo{person}{Zhiwei Qin}, \bibinfo{person}{Fan Zhang}, \bibinfo{person}{Zhaodong Wang}, \bibinfo{person}{Zhe Xu}, \bibinfo{person}{Yintai Ma}, \bibinfo{person}{Hongtu Zhu}, {and} \bibinfo{person}{Jieping Ye}.} \bibinfo{year}{2019}\natexlab{}.
\newblock \showarticletitle{A deep value-network based approach for multi-driver order dispatching}. In \bibinfo{booktitle}{\emph{Proceedings of the 25th ACM SIGKDD international conference on knowledge discovery \& data mining}}. \bibinfo{pages}{1780--1790}.
\newblock


\bibitem[Tang et~al\mbox{.}(2021)]%
        {tang2021value}
\bibfield{author}{\bibinfo{person}{Xiaocheng Tang}, \bibinfo{person}{Fan Zhang}, \bibinfo{person}{Zhiwei Qin}, \bibinfo{person}{Yansheng Wang}, \bibinfo{person}{Dingyuan Shi}, \bibinfo{person}{Bingchen Song}, \bibinfo{person}{Yongxin Tong}, \bibinfo{person}{Hongtu Zhu}, {and} \bibinfo{person}{Jieping Ye}.} \bibinfo{year}{2021}\natexlab{}.
\newblock \showarticletitle{Value function is all you need: A unified learning framework for ride hailing platforms}. In \bibinfo{booktitle}{\emph{Proceedings of the 27th ACM SIGKDD Conference on Knowledge Discovery \& Data Mining}}. \bibinfo{pages}{3605--3615}.
\newblock


\bibitem[Tong et~al\mbox{.}(2021)]%
        {tong2021combinatorial}
\bibfield{author}{\bibinfo{person}{Yongxin Tong}, \bibinfo{person}{Dingyuan Shi}, \bibinfo{person}{Yi Xu}, \bibinfo{person}{Weifeng Lv}, \bibinfo{person}{Zhiwei Qin}, {and} \bibinfo{person}{Xiaocheng Tang}.} \bibinfo{year}{2021}\natexlab{}.
\newblock \showarticletitle{Combinatorial optimization meets reinforcement learning: Effective taxi order dispatching at large-scale}.
\newblock \bibinfo{journal}{\emph{IEEE Transactions on Knowledge and Data Engineering}} (\bibinfo{year}{2021}).
\newblock


\bibitem[Wang et~al\mbox{.}(2019)]%
        {wang2019adaptive}
\bibfield{author}{\bibinfo{person}{Yansheng Wang}, \bibinfo{person}{Yongxin Tong}, \bibinfo{person}{Cheng Long}, \bibinfo{person}{Pan Xu}, \bibinfo{person}{Ke Xu}, {and} \bibinfo{person}{Weifeng Lv}.} \bibinfo{year}{2019}\natexlab{}.
\newblock \showarticletitle{Adaptive dynamic bipartite graph matching: A reinforcement learning approach}. In \bibinfo{booktitle}{\emph{2019 IEEE 35th international conference on data engineering (ICDE)}}. IEEE, \bibinfo{pages}{1478--1489}.
\newblock


\bibitem[Wang et~al\mbox{.}(2018)]%
        {wang2018deep}
\bibfield{author}{\bibinfo{person}{Zhaodong Wang}, \bibinfo{person}{Zhiwei Qin}, \bibinfo{person}{Xiaocheng Tang}, \bibinfo{person}{Jieping Ye}, {and} \bibinfo{person}{Hongtu Zhu}.} \bibinfo{year}{2018}\natexlab{}.
\newblock \showarticletitle{Deep reinforcement learning with knowledge transfer for online rides order dispatching}. In \bibinfo{booktitle}{\emph{2018 IEEE International Conference on Data Mining (ICDM)}}. IEEE, \bibinfo{pages}{617--626}.
\newblock


\bibitem[Xu et~al\mbox{.}(2018)]%
        {xu2018large}
\bibfield{author}{\bibinfo{person}{Zhe Xu}, \bibinfo{person}{Zhixin Li}, \bibinfo{person}{Qingwen Guan}, \bibinfo{person}{Dingshui Zhang}, \bibinfo{person}{Qiang Li}, \bibinfo{person}{Junxiao Nan}, \bibinfo{person}{Chunyang Liu}, \bibinfo{person}{Wei Bian}, {and} \bibinfo{person}{Jieping Ye}.} \bibinfo{year}{2018}\natexlab{}.
\newblock \showarticletitle{Large-scale order dispatch in on-demand ride-hailing platforms: A learning and planning approach}. In \bibinfo{booktitle}{\emph{Proceedings of the 24th ACM SIGKDD International Conference on Knowledge Discovery \& Data Mining}}. \bibinfo{pages}{905--913}.
\newblock


\bibitem[Yang et~al\mbox{.}(2020)]%
        {yang2020optimizing}
\bibfield{author}{\bibinfo{person}{Hai Yang}, \bibinfo{person}{Xiaoran Qin}, \bibinfo{person}{Jintao Ke}, {and} \bibinfo{person}{Jieping Ye}.} \bibinfo{year}{2020}\natexlab{}.
\newblock \showarticletitle{Optimizing matching time interval and matching radius in on-demand ride-sourcing markets}.
\newblock \bibinfo{journal}{\emph{Transportation Research Part B: Methodological}}  \bibinfo{volume}{131} (\bibinfo{year}{2020}), \bibinfo{pages}{84--105}.
\newblock


\bibitem[Zeng et~al\mbox{.}(2023)]%
        {zeng2023deep}
\bibfield{author}{\bibinfo{person}{Hao Zeng}, \bibinfo{person}{Qiong Wu}, \bibinfo{person}{Kunpeng Han}, \bibinfo{person}{Junying He}, {and} \bibinfo{person}{Haoyuan Hu}.} \bibinfo{year}{2023}\natexlab{}.
\newblock \showarticletitle{A Deep Reinforcement Learning Approach for Online Parcel Assignment}. In \bibinfo{booktitle}{\emph{Proceedings of the 2023 International Conference on Autonomous Agents and Multiagent Systems}}. \bibinfo{pages}{1961--1968}.
\newblock


\bibitem[Zeng et~al\mbox{.}(2018)]%
        {zeng2018latency}
\bibfield{author}{\bibinfo{person}{Yuxiang Zeng}, \bibinfo{person}{Yongxin Tong}, \bibinfo{person}{Lei Chen}, {and} \bibinfo{person}{Zimu Zhou}.} \bibinfo{year}{2018}\natexlab{}.
\newblock \showarticletitle{Latency-oriented task completion via spatial crowdsourcing}. In \bibinfo{booktitle}{\emph{2018 IEEE 34th International Conference on Data Engineering (ICDE)}}. IEEE, \bibinfo{pages}{317--328}.
\newblock


\bibitem[Zhou et~al\mbox{.}(2019)]%
        {zhou2019multi}
\bibfield{author}{\bibinfo{person}{Ming Zhou}, \bibinfo{person}{Jiarui Jin}, \bibinfo{person}{Weinan Zhang}, \bibinfo{person}{Zhiwei Qin}, \bibinfo{person}{Yan Jiao}, \bibinfo{person}{Chenxi Wang}, \bibinfo{person}{Guobin Wu}, \bibinfo{person}{Yong Yu}, {and} \bibinfo{person}{Jieping Ye}.} \bibinfo{year}{2019}\natexlab{}.
\newblock \showarticletitle{Multi-agent reinforcement learning for order-dispatching via order-vehicle distribution matching}. In \bibinfo{booktitle}{\emph{Proceedings of the 28th ACM International Conference on Information and Knowledge Management}}. \bibinfo{pages}{2645--2653}.
\newblock


\end{thebibliography}


\end{document}